\documentclass[10pt,twocolumn,letterpaper]{article}

\usepackage{wacv}              

\newcommand{\scoreci}[2]{\shortstack{#1\\[-1pt]{\scriptsize $\pm\,#2$}}}

\usepackage{amsmath}
\usepackage{booktabs}
\usepackage{microtype}
\usepackage{placeins}

\definecolor{wacvblue}{rgb}{0.21,0.49,0.74}
\usepackage[pagebackref,breaklinks,colorlinks,allcolors=wacvblue]{hyperref}

\def\wacvPaperID{1043} 
\def\confName{WACV}
\def\confYear{2027}

\title{Reuse, Don’t Retrain: Efficient, Modality-Versatile Geospatial Foundation Models through Feature Distillation}

\author{Mario Hamon\thanks{Equal contribution.}
\qquad
Pierre Adorni*
\qquad
Minh-Tan Pham\\
\\
IRISA\\
Université Bretagne Sud, UMR 6074 \\
{\tt\small first-name.last-name@irisa.fr}
}

\begin{document}
\maketitle
\begin{abstract}

Geospatial foundation models (GeoFMs) are costly to pretrain, yet model families often repeat this process across backbone scales and sensing modalities. We investigate feature distillation as a more efficient way to reuse existing representations. First, we distill frozen CROMA and TerraMind teachers into smaller Vision Transformers through patch-level feature matching. Across nine PANGAEA tasks, distilled TerraMind Tiny, Small, and Base outperform their scale-matched released counterparts by 2.76, 3.02, and 1.87 mean IoU points, respectively; each also matches or exceeds the next-larger released model. Distilled CROMA-Base gains 2.05 points. TerraMind-Base required only 43 A6000 GPU-hours, approximately 0.5\% of the reported cost of independent pretraining. Second, we introduce TEDI, an unpaired multi-expert approach combining TerraMind for multispectral imagery and DINOv3 for high-resolution RGB. Separate patch embeddings route unpaired inputs into a shared Transformer trunk, removing the need for co-registered data. TEDI-Large achieves the highest single-model mean in our PANGAEA-9 comparison (62.51 mIoU), outperforming both teachers overall while retaining their modality-specific strengths. TEDI-Base reaches 61.75 mIoU with 72\% fewer parameters. These results show that distillation improves the GeoFM accuracy-compute frontier while consolidating complementary experts into a single versatile backbone.
\end{abstract}

\begin{figure*}[t]
    \centering
    \includegraphics[width=\linewidth]{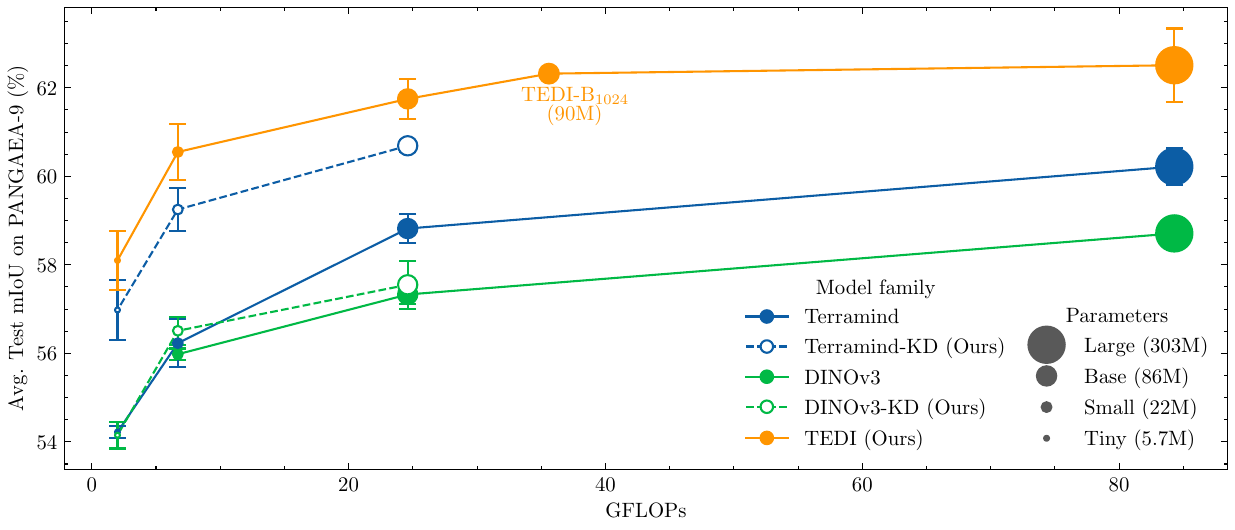}
    \caption{\textbf{Accuracy--compute trade-off.} Released TerraMind models,
    their single-teacher distilled counterparts (TerraMind-KD), and
    multi-expert TEDI students are evaluated on the same nine PANGAEA tasks.
    Marker area indicates parameter count. Distillation shifts the entire
    model family toward a better trade-off; adding a complementary HR expert
    further improves average performance. TEDI-B$_{1024}$ reallocates Base
    capacity toward a wider token embedding. Final comparisons use test
    scores; error bars show 3-run standard deviation.}
    \label{fig:pareto}
\end{figure*}
    
\section{Introduction}

Geospatial foundation models (GeoFMs) have made it possible to adapt a common
pretrained encoder to many Earth-observation tasks with limited annotations.
Their practical use, however, is constrained by two forms of redundancy.
First, large encoders are costly to apply to continent-scale image archives,
which motivates the release of several backbone sizes. Second, Earth
observation spans heterogeneous sensors and resolutions, which often leads to
separate models for multispectral satellite imagery, synthetic-aperture radar,
and high-resolution (HR) optical imagery. A deployment that must cover both
model scales and modalities may therefore require many expensive pretraining
runs and several independent checkpoints.

Computer-vision foundation model families increasingly avoid the first source
of redundancy. Rather than repeat self-supervised pretraining for every
backbone, they train a strong model once and distill its representations into
smaller students~\cite{oquab2023dinov2,simeoni2026dinov,ren2023tinymim}.
GeoFM families commonly follow the opposite practice. CROMA~\cite{fuller2023croma}
and TerraMind~\cite{jakubik2025terramind}, for example, provide multiple sizes
obtained through scale-specific pretraining. Whether this additional
pretraining is necessary has not been established through a systematic
comparison.

The modality problem is related. Existing GeoFMs contain complementary
knowledge: a multispectral model can exploit spectral signatures unavailable
to an RGB model, while a large vision foundation model trained on HR imagery
can provide stronger fine-grained spatial features. Multi-teacher
distillation has aggregated such complementary representations in general
computer vision~\cite{ranzinger2024amradio}, but its usual formulation assumes
that every teacher consumes the same input. In Earth observation, modality
experts may instead require different channels, resolutions, and datasets.

We argue that both inefficiencies can be addressed by the same operation:
\emph{reuse strong pretrained representations instead of recreating them}.
We investigate a straightforward recipe in which a student matches the
patch representations of frozen teachers through lightweight projection
heads. With one teacher, this derives a family of efficient students. With
multiple experts, modality-specific input and output adapters route independent
training samples to a shared Transformer trunk. The training images do not need to
be paired, co-registered, or even drawn from the same corpus.

Our experiments use CROMA and TerraMind to study model compression, then combine
TerraMind with DINOv3~\cite{simeoni2026dinov} in a modality-versatile family
called TEDI (\textbf{TE}rraMind--\textbf{DI}NOv3). The same progression makes
the two settings parts of one question: how far can already-trained GeoFMs be
reused across both scale and modality?

Our contributions are:
\begin{enumerate}
    \item We provide a comparative study of feature distillation for deriving
    smaller GeoFM backbones. Distilled TerraMind Tiny, Small, and Base models
    consistently outperform their scale-matched released counterparts and
    establish a stronger accuracy--compute frontier.
    \item We extend the same formulation to \emph{unpaired multi-expert
    distillation}. TEDI combines two independently pretrained and frozen
    modality experts using modality-specific patch embeddings and projection
    heads around a shared student trunk.
    \item We evaluate compression and aggregation under a nine-task benchmark, 
    analyze teacher multiplicity, training extent, data diversity,
    and token-width allocation, and report the resulting deployment and
    distillation costs.
\end{enumerate}

\begin{figure*}[t]
    \centering
    \includegraphics[width=\linewidth]{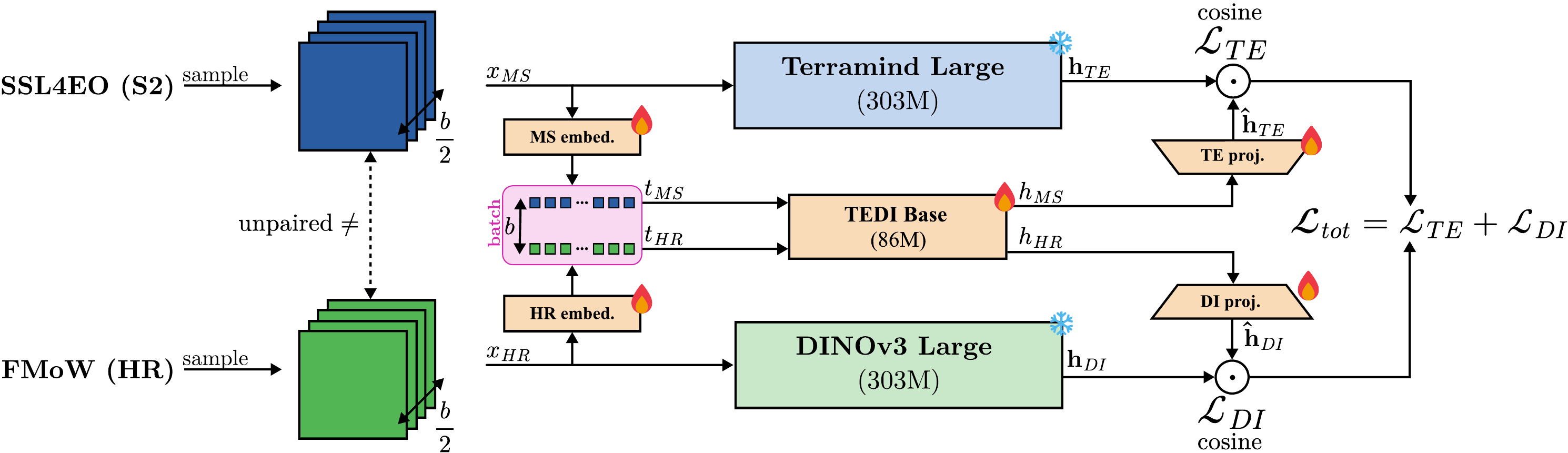}
    \caption{\textbf{TEDI training.} Independent Sentinel-2 and HR samples are
    routed through modality-specific patch embeddings and a shared student
    trunk. Each sample is matched only to its corresponding frozen pretrained
    expert through a training-only projection head. Unlike a paired
    cross-modal objective, TEDI requires neither spatial correspondence nor a
    shared input representation across experts.}
    \label{fig:tedi}
\end{figure*}
\section{Related Work}

\subsection{Geospatial foundation models}

GeoFMs learn transferable representations from large unlabeled
Earth-observation archives. CROMA combines radar--optical contrastive learning
with masked reconstruction~\cite{fuller2023croma}; TerraMind scales
multimodal generative pretraining to nine modalities and releases a family of
Vision Transformer (ViT) backbones~\cite{jakubik2025terramind}. Other work
adapts general vision models through continual pretraining
\cite{mendieta2023gfm} or constructs flexible encoders for heterogeneous
sensors. AnySat uses scale-adaptive encoders and joint-embedding predictive
learning across many resolutions and modalities~\cite{astruc2025anysat},
whereas SkySense V2 shares a Transformer using modality prompts and adaptive
patch merging~\cite{zhang2025skysensev2}. Copernicus-FM uses dynamic
hypernetworks and metadata conditioning to support diverse Sentinel
modalities~\cite{wang2025copernicus}.

These approaches demonstrate the value of broad pretraining, but typically
train the versatile backbone directly. Our question is complementary: \textit{once
strong specialist checkpoints exist, can they be compressed and consolidated
without repeating their original objectives?}

\subsection{Knowledge distillation and model reuse}

Knowledge distillation originally transferred predictive distributions from
a cumbersome teacher or ensemble into a deployable student
\cite{hinton2015distilling}. For ViTs and self-supervised models, recent work
often aligns hidden representations rather than class logits. ViTKD studies
which Transformer features to transfer~\cite{yang2022vitkd}, while TinyMIM compares feature and relation distillation for
masked-image-modeling teachers~\cite{ren2023tinymim}. DINOv2 and DINOv3 use
large teachers to derive smaller members of their released model families
\cite{oquab2023dinov2,simeoni2026dinov}.

Distillation can also consolidate teachers with different capabilities.
AM-RADIO matches representations from several vision foundation models in one
input, whereas ours specialize in different sensor domains and can be trained
using independent modality-specific datasets.

DEO~\cite{wolf2026deo} is concurrent and related in its goal of producing one
optical/multispectral encoder, but it differs in the source of supervision. DEO
combines a DINO-style multispectral \emph{self-distillation} teacher (an
exponential-moving-average copy of the student) with one frozen optical VFM.
It therefore augments multispectral self-supervised pretraining with external
DINOv3 knowledge. TEDI instead performs model reuse: both TerraMind and
DINOv3 are independently pretrained, frozen external experts, and no
self-distillation objective or paired cross-modal sample is required.

\subsection{GeoFM evaluation}

GeoFM evaluation has increasingly moved from isolated downstream tasks toward
standardized, multi-task benchmarks. GEO-Bench introduced a curated suite of
six classification and six segmentation tasks, together with consistent
evaluation and aggregate-reporting procedures~\cite{lacoste2023geobench}.
PANGAEA broadened this scope toward geographically diverse, application-oriented
tasks spanning different sensors, resolutions, and temporal settings, while also
evaluating supervised baselines and limited-label regimes
\cite{marsocci2025pangaea}. More recently, GEO-Bench-2 expanded the task coverage
to 19 permissively licensed datasets, including classification, segmentation,
regression, object detection, and instance segmentation, and organized them into
overlapping capability groups reflecting properties such as modality,
spectral coverage, resolution, and temporality~\cite{simumba2025geobench2}.
Complementary benchmarks examine more specific aspects of transferability; REOBench evaluates the robustness of GeoFMs to appearance and geometric
corruptions~\cite{li2025reobench}. Collectively, these efforts show that model
rankings depend strongly on the evaluated capabilities and adaptation protocol,
rather than being captured by a single universally representative task.

We adopt PANGAEA because its frozen-backbone protocol and heterogeneous tasks
are well suited to evaluating whether TEDI preserves complementary
modality-specific representations. We use one fixed nine-task subset for every
reproduced model and report three downstream runs. For broader context, we also
recompute PANGAEA-9 means from published per-task results for DOFA, EoS-FM, and
standard baselines; these reported entries are kept separate from our controlled
reruns.
\section{GeoFM Reuse through Feature Distillation}
\label{sec:method}

Let $\mathcal{M}$ denote the set of modalities, with a frozen teacher $T_m$ associated with each modality $m\in\mathcal{M}$. An image $x_m$ is processed by both $T_m$ and a shared student Transformer $S_\theta$. The student uses a modality-specific patch embedding $E_m$ to accommodate different input channel counts and a training-only projection head $P_m$ to match the corresponding teacher dimension. We define the projected student and teacher targets as
\begin{align}
    \mathbf{\hat{h}}_{m} &= P_m\!\left(S_\theta(E_m(x_m))\right),\\
    \mathbf{h}_{m} &= T_m(x_m).
\end{align}
We minimize the mean patch-level cosine distance
\begin{equation}
    \mathcal{L}_{m}
    =\frac{1}{N_m}\sum_{i=1}^{N_m}
    \left[1-\cos\left(\mathbf{\hat{h}}_{m,i},\mathbf{h}_{m,i}\right)\right],
    \label{eq:distill}
\end{equation}

where $N_m$ is the number of patch tokens, and $\mathbf{h}_{m,i}$ corresponds to the $i$-th token of the feature map $\mathbf{h}_{m}$. The full objective is
\begin{equation}
    \mathcal{L}=\sum_{m\in\mathcal{M}}\lambda_m\mathcal{L}_m.
    \label{eq:multi}
\end{equation}
Projection heads are discarded after training. The shared trunk and required
patch embeddings are retained.
Class tokens are excluded in every experiment. The student architecture produces none, as it is designed for downstream dense, patch-level tasks rather than global image-level summarization. Each projection head $P_m$ is a two-layer GELU MLP (linear--GELU--linear) with default initialization. Teacher features are used without explicit normalization and are learned by
the student in their native scale; the cosine formulation of $\mathcal{L}_m$
nonetheless normalizes each patch representation implicitly.

\subsection{Deriving an efficient model family}

With $|\mathcal{M}|=1$, \crefrange{eq:distill}{eq:multi} reduce to
single-teacher feature distillation. We study two teacher families to separate
the general effect from a particular pretraining recipe.

CROMA-Large is distilled into the same ViT-Base architecture as the released
CROMA-Base model, including its positional-bias implementation.
Both models are trained from SSL4EO-S12, making this an
architecture- and corpus-matched comparison between the released pretraining
route and our distillation route. 

TerraMind-Large is
distilled into ViT-Base, ViT-Small, and ViT-Tiny students. These students use
rotary positional embeddings (RoPE)~\cite{su2024roformer} and standard GELU feed-forward blocks. In other
words, they are not exact reduced copies of TerraMind: positional information
is introduced by rotating queries and keys inside self-attention, and each
feed-forward sublayer is a conventional linear--GELU--linear MLP, replacing
TerraMind's native positional-embedding and gated-MLP implementations. The
TerraMind comparison therefore evaluates this complete student design in
addition to the distillation procedure.

\subsection{Unpaired multi-expert distillation}

With multiple elements in $\mathcal{M}$, each mini-batch contains
modality-specific samples and ~\cref{eq:multi} aggregates their losses. The
inputs do not need cross-modal correspondence: a Sentinel-2 image is evaluated
only by the Sentinel-2 expert, and an HR image only by the HR expert. Shared
student parameters nevertheless receive gradients from both branches.

The general pretraining scheme is described in \cref{fig:tedi}. TEDI uses frozen TerraMind-Large and DINOv3-Large teachers. Each training batch contains
an equal amount of images from each modality. Two patch embeddings handle
the distinct channel configurations, and two two-layer GELU MLP
projection heads map student patch features into the respective teacher
spaces. All Transformer blocks are
shared. We train Tiny, Small, Base, and Large variants. Standard TEDI-Base
uses a 768-dimensional embedding, 12 blocks, and an MLP expansion ratio of 4. 
All TEDI backbones use RoPE to encode positional information.
We separately study TEDI-B$_{1024}$, which retains 12 blocks but widens the
embedding to 1024 and reduces the MLP ratio to 1.5
(\cref{sec:width_ablation}).

\begin{table}[t]
    \centering
    \caption{\textbf{Context against PANGAEA baselines.} Mean test mIoU on
    PANGAEA-9. DOFA, EoS-FM and DEO are modality-versatile published baselines.
    Reported means are recomputed from published per-task values after applying
    our fixed task subset and have no uncertainty estimates. Our entries show
    one sample standard deviation across three runs. \textsuperscript{\dag} 
    means fine-tuned, otherwise backbones are frozen.}
    \label{tab:pangaea9_context}
    \setlength{\tabcolsep}{4pt}
    \begin{tabular}{lrrl}
        \toprule
        \textbf{Model} & \textbf{Params.} & \textbf{Mean} & \textbf{Source} \\
        \midrule
        U-Net~\cite{marsocci2025pangaea}\textsuperscript{\dag} & ${\sim}8$M & 58.82 & reported \\
        ViT-B~\cite{marsocci2025pangaea}\textsuperscript{\dag} & 87M & 55.61 & reported \\
        DOFA~\cite{xiong2024neural} & 112M & 54.51 & reported \\
        EoS-FM~\cite{adorni2026eosfm} & 76M & 60.18 & reported \\
        DEO~\cite{wolf2026deo} & 87M & \scoreci{58.07}{0.22} & reproduced \\
        \midrule
        CROMA-Large & 303M & \scoreci{57.57}{0.09} & reproduced \\
        DINOv3-Large & 303M & \scoreci{58.71}{0.36} & reproduced \\
        TerraMind-Large & 303M & \scoreci{60.22}{0.41} & reproduced \\
        \midrule
        CROMA-KD-Base & 86M & \scoreci{57.88}{0.32} & ours \\
        TerraMind-KD-Small & 22M & \scoreci{59.25}{0.49} & ours \\
        TerraMind-KD-Base & 86M & \scoreci{60.69}{0.13} & ours \\
        TEDI-Small & 22M & \scoreci{60.55}{0.64} & ours \\
        TEDI-Base & 86M & \scoreci{61.75}{0.45} & ours \\
        TEDI-B$_{1024}$ & 90M & \scoreci{62.32}{0.13} & ours \\
        TEDI-Large & 303M & \scoreci{\textbf{62.51}}{0.83} & ours \\
        \bottomrule
    \end{tabular}
\end{table}
\section{Experiments}
\label{sec:experiments}

\subsection{Distillation protocol}

Unless specified otherwise, teachers remain frozen and students are trained
on $224\times224$ images. Single-teacher experiments use a batch size of 64,
with SSL4EO-S12~\cite{wang2023ssl4eo} for CROMA and TerraMind, and 
FMoW~\cite{christie2018fmow} for DINOv3. TEDI batches contain equal
numbers of SSL4EO-S12 Sentinel-2 and FMoW HR images, resulting
in a batch size of 128. For SSL4EO-S12 images, we apply a deterministic bilinear-interpolation resize
to $224\times224$, while fMoW images are pre-tiled offline into non-overlapping
$224\times224$ crops. Each active teacher is weighted equally, with $\lambda_m = 1/|\mathcal{M}|$. Training uses AdamW with an initial learning rate of $10^{-4}$, a weight
decay of $10^{-4}$, and a cosine schedule with no warmup. All models are trained in mixed precision on a single GPU, with gradients clipped to a maximum norm of 1.0. As our datasets
are unpaired and do not contain the same number of images, we preserve a
comparable total update budget rather than a comparable number of dataset
passes, so all students are trained with the same number of steps,
equivalent to 20 epochs over SSL4EO-S12. Validation performance selects a
checkpoint equivalent to 15 epochs over SSL4EO-S12 for downstream test
evaluation. 

\subsection{Evaluation protocol}

We evaluate frozen pretrained encoders with the PANGAEA adaptation
protocol~\cite{marsocci2025pangaea}, using the released code and default
configuration parameters. For every model, the backbone is frozen and a
512-channel UPerNet decoder is trained for 80 epochs with a batch size of 8
on the full labeled training split. We retain the task-specific losses,
preprocessing, AdamW optimizer, learning-rate schedule, and other settings
defined by PANGAEA. The original benchmark contains 11 tasks. We report a
fixed nine-task subset, denoted \emph{PANGAEA-9}. We exclude
BioMassters~\cite{nascetti2023biomassters} because we could not obtain
sufficiently reproducible results for consistent model comparison. This issue
has also affected prior evaluations: TerraMind and THOR omit BioMassters (and
xView2), with TerraMind reporting that the decision followed guidance from the
PANGAEA authors~\cite{jakubik2025terramind,forgaard2026thor}. We retain xView2,
for which our runs were stable, but exclude Crop Type Mapping--South Sudan
\cite{m2019semantic} because high inter-run variability obscured differences
among models. These exclusions are applied identically to every encoder. The
retained tasks comprise five multispectral (MS) evaluations: HLS Burn Scars,
MADOS, PASTIS, Sen1Floods11, and AI4SmallFarms, and four HR
evaluations: xView2, Five Billion Pixels, DynamicEarthNet, and SpaceNet7. All
subsequent references to PANGAEA-9 denote exactly this fixed subset.

\noindent\textbf{Reporting convention.}
Validation scores are used for training decisions and exploratory analyses,
including checkpoint selection, training extent, and data-fraction studies.
Final model comparisons and the prespecified teacher-multiplicity ablation
report test scores after configurations are fixed. Unless stated otherwise,
each downstream result is the mean of three runs on our
hardware. We rerun all teachers and students used in our primary comparisons
under the same codebase because our reproductions of published CROMA and
TerraMind values were slightly but consistently lower than the reported
numbers. \cref{tab:pangaea9_context} also provides broader context using
published baselines. For our three-run evaluations, we report the mean and 
standard deviation. Unet uses its own decoder, all other models follow the 
PANGAEA protocol, using a UPerNet decoder on top of the frozen encoder.


\subsection{Efficient model-family derivation}

\cref{tab:terramind_family} compares the released TerraMind family with
students distilled directly from TerraMind-Large. At every matched scale,
distillation improves mean test IoU: $+1.87$ for Base, $+3.02$ for Small, and
$+2.76$ for Tiny. More importantly for deployment, every student reaches or
exceeds the next-larger released model. TerraMind-KD-Base slightly surpasses
TerraMind-Large, TerraMind-KD-Small surpasses TerraMind-Base, and
TerraMind-KD-Tiny surpasses TerraMind-Small. \cref{fig:pareto} shows that
the improvement is consistent across the accuracy--compute frontier.
\begin{table}[t]
    \centering
    \caption{\textbf{Single-teacher model-family derivation.} Mean test mIoU
    on PANGAEA-9 across three downstream runs, with one sample
    standard deviation shown below the mean. Gains compare matched backbone
    sizes. CROMA results use its Sentinel-2 encoder.}
    \label{tab:terramind_family}
    \setlength{\tabcolsep}{3.5pt}
    \begin{tabular}{llrrr}
        \toprule
        \textbf{Family} & \textbf{Backbone} & \textbf{Released} &
        \textbf{KD} & \textbf{Gain} \\
        \midrule
        TerraMind & Large & \scoreci{60.22}{0.41} & -- & -- \\
        & Base  & \scoreci{58.82}{0.32} &
        \scoreci{\textbf{60.69}}{0.13} & $+1.87$ \\
        & Small & \scoreci{56.23}{0.54} &
        \scoreci{\textbf{59.25}}{0.49} & $+3.02$ \\
        & Tiny  & \scoreci{54.22}{0.13} &
        \scoreci{\textbf{56.98}}{0.67} & $+2.76$ \\
        \midrule
        CROMA & Large & \scoreci{57.57}{0.09} & -- & -- \\
        & Base & \scoreci{55.83}{0.66} &
        \scoreci{\textbf{57.88}}{0.32} & $+2.05$ \\
        \bottomrule
    \end{tabular}
\end{table}
CROMA provides the better-controlled comparison: CROMA-KD-Base uses the released Base architecture and SSL4EO-S12 corpus, and CROMA uses a separate unimodal optical encoder rather than a shared multimodal trunk. It improves over the released CROMA-Base by 2.05 points after 15 distillation epoch-equivalents rather than the released Base model's 300 pretraining epochs; a 20-times reduction in training length. Thus, for CROMA, the practical reuse route produces a stronger Base model without an architectural or corpus advantage. TerraMind’s gains across three scales are consistent with this result, but could also reflect its different student design (RoPE and standard GELU MLPs, Sentinel-2 uni-modal). \cref{fig:pareto} also shows that our distilled DINOv3-KD model family matches rather than improves the original DINOv3 family, which was already distilled from its largest checkpoint.

\subsection{Unpaired multi-expert aggregation}

TerraMind and DINOv3 are complementary experts under PANGAEA-9:
TerraMind-Large is stronger on the MS mean, while DINOv3-Large is stronger on
the HR mean. \cref{tab:tedi_results} evaluates whether
TEDI can retain both capabilities in one Base student.

\begin{table*}[t]
    \centering
    \caption{\textbf{Multi-expert aggregation.} Mean test mIoU on the nine
    PANGAEA-9 tasks. MS and HR average their respective five and four tasks, All over all nine. The expert oracle selects the better teacher
    independently for each task and therefore requires two deployed models.
    Results are averaged across three downstream runs, with one sample
    standard deviation shown below the mean. HLS=HLSBurnScars, MAD=Mados, PAS=PASTIS, S1F=Sen1Floods11, AI4=AI4SmallFarms, xV2=xView2, FBP=FiveBillionPixels, DEN=DynamicEarthNet, SN7=SpaceNet7.}
    \label{tab:tedi_results}
    \setlength{\tabcolsep}{3.2pt}
    \begin{tabular}{l|rrrrr|rrrr|rrr}
        \toprule
        \textbf{Model} &
        \multicolumn{5}{c|}{\textbf{Multispectral}} &
        \multicolumn{4}{c|}{\textbf{High resolution}} &
        \textbf{MS} & \textbf{HR} & \textbf{All} \\
        & HLS & MAD & PAS & S1F & AI4 & xV2 & FBP & DEN & SN7
        & & & \\
        \midrule
        TerraMind-L & \scoreci{82.41}{1.12} &
        \scoreci{\textbf{70.88}}{1.83} &
        \scoreci{\textbf{42.68}}{0.02} & \scoreci{90.62}{0.22} &
        \scoreci{39.59}{0.43} & \scoreci{56.54}{0.48} &
        \scoreci{62.34}{0.19} & \scoreci{36.74}{0.90} &
        \scoreci{60.20}{0.26} & \scoreci{65.24}{0.66} &
        \scoreci{53.95}{0.09} & \scoreci{60.22}{0.41} \\
        DINOv3-L & \scoreci{78.57}{0.42} &
        \scoreci{66.57}{3.54} & \scoreci{25.83}{0.39} &
        \scoreci{75.58}{0.66} & \scoreci{\textbf{45.13}}{0.18} &
        \scoreci{\textbf{61.38}}{0.71} &
        \scoreci{\textbf{74.71}}{0.05} & \scoreci{37.58}{0.82} &
        \scoreci{63.05}{0.58} & \scoreci{58.34}{0.82} &
        \scoreci{\textbf{59.18}}{0.22} & \scoreci{58.71}{0.36} \\
        \midrule
        TEDI-L & \scoreci{\textbf{83.64}}{0.40} &
        \scoreci{69.36}{7.47} & \scoreci{41.60}{0.41} &
        \scoreci{\textbf{90.97}}{0.34} & \scoreci{40.86}{0.58} &
        \scoreci{61.32}{1.26} & \scoreci{72.57}{0.58} &
        \scoreci{38.94}{0.93} & \scoreci{\textbf{63.34}}{0.12} &
        \scoreci{\textbf{65.29}}{1.47} & \scoreci{59.04}{0.06} &
        \scoreci{\textbf{62.51}}{0.83} \\
        TEDI-B & \scoreci{83.12}{0.36} &
        \scoreci{66.96}{3.47} & \scoreci{40.04}{0.65} &
        \scoreci{90.82}{0.30} & \scoreci{39.66}{0.52} &
        \scoreci{60.90}{0.63} & \scoreci{71.42}{0.33} &
        \scoreci{\textbf{39.49}}{0.58} &
        \scoreci{\textbf{63.34}}{0.31} &
        \scoreci{64.12}{0.79} & \scoreci{58.79}{0.27} &
        \scoreci{\textbf{61.75}}{0.45} \\
        \midrule
        Expert oracle & 82.41 & 70.88 & 42.68 & 90.62 & 45.13 &
        61.38 & 74.71 & 37.58 & 63.05 & 66.34 & 59.18 & 63.16 \\
        \bottomrule
    \end{tabular}
\end{table*}

TEDI-Large obtains the highest single-model average, $62.51\pm0.83$, while
matching TerraMind on the MS mean within 0.05 points and remaining 0.14 below
DINOv3 on the HR mean. It remains 0.65 points below the task-wise expert
oracle. TEDI-Base obtains $61.75\pm0.45$, only 0.76 points lower despite using
72\% fewer parameters and 71\% fewer standardized GFLOPs than TEDI-Large. The
three runs do not support a precise estimate of this scaling gain, but the
operating-point comparison is clear: Large nearly retains both specialist
domain means, whereas Base gives up less than one mean point for a much smaller
backbone. \cref{sec:width_ablation} examines TEDI-B$_{1024}$, an intermediate design
that increases token width without adopting ViT-Large depth.

\subsection{System cost}

The central application claim concerns the complete deployment trade-off, not
only accuracy. \cref{tab:resources} summarizes the TEDI student scales.
Projection heads are removed after distillation, so inference retains only the
shared backbone and the two input embeddings. Parameter and storage counts
include both retained embeddings, although only the selected embedding is
evaluated for any given image. Storage values assume four-byte FP32 weights and
exclude serialization overhead.
\begin{table*}[t]
    \centering
    \caption{\textbf{TEDI model and distillation resources.} Parameter and
    storage counts include the shared trunk and both retained input embeddings;
    training-only heads are excluded. GFLOPs use \texttt{ptflops} with a
    standardized 10-channel $224\times224$ input. FP32 batch-1 inference uses
    an RTX 2080 Ti after 20 warm-up and 100 synchronized timed runs; peak memory
    includes resident weights. Distillation times are elapsed hours for the
    corresponding TEDI single-GPU runs and are not normalized across hardware.}
    \label{tab:resources}
    \small
    \setlength{\tabcolsep}{4pt}
    \begin{tabular}{lrrrrrrr}
        \toprule
        \textbf{Model} & \textbf{Params.} & \textbf{GFLOPs} &
        \textbf{FP32} & \textbf{Images/s} &
        \textbf{Latency p50/p95 (ms)} & \textbf{Peak MB} &
        \textbf{Distill h / GPU} \\
        \midrule
        TEDI-T/16 & 5.7M & 2.0  & 23MB   & 165.02 & 6.07 / 6.43 & 34.92 & 56.29 / V100 \\
        TEDI-S/16 & 22M  & 6.7  & 88MB   & 145.96 & 7.33 / 7.60 & 101.06 & 91.07 / Titan RTX \\
        TEDI-B/16 & 86M  & 24.6 & 344MB  & 86.62 & 12.88 / 13.24 & 353.12 & 63.75 / A6000 \\
        TEDI-B$_{1024}$/16 & 90M & 35.6 & 360MB & 73.44 & 13.36 / 16.08 & 369.02 & 69.40 / A6000 \\
        TEDI-L/16 & 303M & 84.3 & 1.21GB & 30.86 & 38.80 / 39.76 & 1187.27 & 114.77 / A6000 \\
        \bottomrule
    \end{tabular}
\end{table*}

Under this standardized FLOP setting, TEDI-Base contains 86M parameters and
requires 24.6 GFLOPs, compared with 303M parameters and 84.3 GFLOPs for a
Large teacher. Deploying both Large experts stores
606M parameters (approximately 2.42GB in FP32), although each image uses only
one. TEDI stores 344MB, reducing model storage by 86\% and routed FLOPs by 71\%.
Relative to the two Base students in \cref{tab:teacher_ablation}, TEDI
halves backbone storage while preserving routed-backbone FLOPs.
In the measured inference benchmark, the larger TEDI-B$_{1024}$ variant
processes 73.44 images/s with 13.36 ms median batch-1 latency and 369MB peak
memory. Relative to TEDI-Large,
this is $2.38\times$ higher throughput, 66\% lower median latency, and 69\%
lower peak memory, while preserving most of the performance.

The training footprint also illustrates the practical cost of the distillation
route. TerraMind reports 9,216 A100 GPU-hours to pretrain
TerraMind-Base~\cite{jakubik2025terramind}, whereas the two-expert TEDI-Base
run required 63.75 hours on one RTX A6000: roughly 145 times fewer reported
accelerator-hours. This is a marginal-cost comparison against reusing existing
teachers, not a hardware-normalized training-efficiency experiment. Even with
that qualification, the two-order-of-magnitude gap is strong evidence that
multi-expert distillation is far cheaper than repeating full pretraining.

\section{Analysis and Ablations}
\label{sec:ablations}

\subsection{Training extent and data diversity}

\begin{figure*}[t]
    \centering
    \begin{minipage}[t]{0.48\linewidth}
        \centering
        \includegraphics[width=1\linewidth]{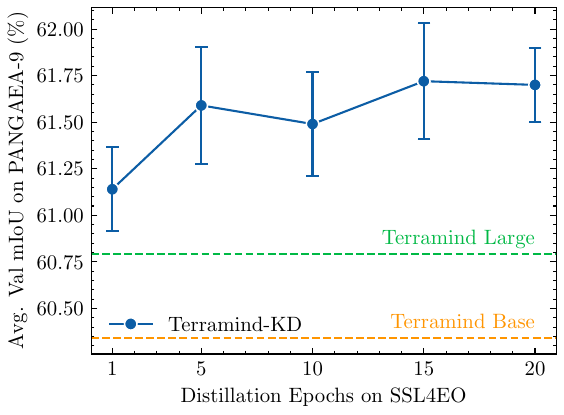}
    \end{minipage}
    \hfill
    \begin{minipage}[t]{0.48\linewidth}
        \centering
        \includegraphics[width=1\linewidth]{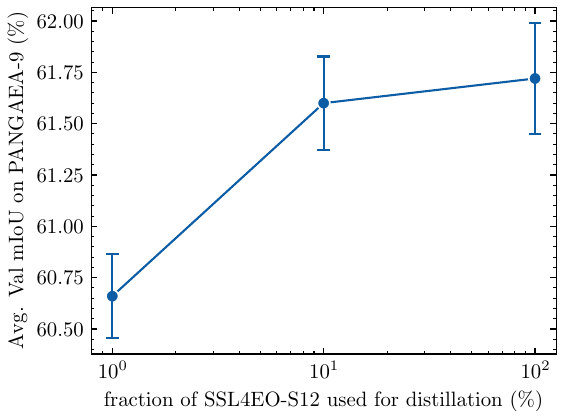}
    \end{minipage}
    \caption{\textbf{Validation analyses.} TerraMind-KD-Base val performance vs. distillation extent (left) and fraction of unique SSL4EO-S12 locations
    (right). Data-fraction runs hold update count fixed. Error bars show one sample standard deviation across three downstream runs.}
    \label{fig:ablation_curves}
\end{figure*}

Mean validation performance rises from approximately 61.1 at epoch 1 to 61.7
at epoch 15 and is nearly unchanged at epoch 20
(\cref{fig:ablation_curves}, left). From epoch 5 onward, means differ by at
most 0.2 points relative to comparable run-to-run variability. The curve does
not identify a sharp optimum; we select epoch 15 because it has the highest
validation mean, before accessing test results.

Holding updates fixed, validation increases from approximately 60.7 with 1\%
of SSL4EO-S12 to 61.6 with 10\% and 61.7 with 100\%
(\cref{fig:ablation_curves}, right). Thus 1\% is insufficient, whereas
three runs do not resolve a difference between 10\% and 100\%. Diversity
matters at least when the corpus is reduced severely. SSL4EO-S12 is itself one of the source datasets used to pretrain TerraMind, alongside the substantially 
broader MajorTOM-Core coverage and
additional derived modalities~\cite{jakubik2025terramind}. Our distillation
data are therefore drawn from a geographically narrower subset of the
teacher's original data distribution, leaving open whether using more of
TerraMesh would improve the student further.

\subsection{What does the second expert contribute?}

TEDI can improve the overall average either through positive transfer from a
second expert or merely because it sees an additional image corpus. Conversely,
sharing a finite-capacity trunk may degrade each specialist domain. We compare TerraMind-KD-Base, DINOv3-KD-Base, and TEDI-Base using the same
ViT-Base trunk, optimization schedule, per-modality batch size, and update
count. Single-expert updates use 64 images; TEDI uses 64 from each corpus, 128
total. The per-modality budget is therefore fixed, but TEDI processes twice as
many images per update. This tests the practical joint addition of a second
teacher and its corpus, including any interference from sharing capacity.

\begin{table}[t]
    \centering
    \caption{\textbf{Teacher-multiplicity ablation.} Avg. test mIoU across
    three downstream runs, with standard deviation. Students use the same architecture and optimization budget. Two-student deployment: MS tasks to TerraMind-KD and HR tasks to DINOv3-KD.}
    \label{tab:teacher_ablation}
    \setlength{\tabcolsep}{3.5pt}
    \begin{tabular}{lrrrr}
        \toprule
        \textbf{Student} & \textbf{MS} & \textbf{HR} & \textbf{All} &
        \textbf{Params.} \\
        \midrule
        TerraMind-KD-B & \scoreci{\textbf{65.41}}{0.14} &
        \scoreci{54.80}{0.13} & \scoreci{60.69}{0.13} & 86M \\
        DINOv3-KD-B & \scoreci{57.21}{0.49} &
        \scoreci{57.98}{0.89} & \scoreci{57.55}{0.54} & 86M \\
        TEDI-B & \scoreci{64.12}{0.79} &
        \scoreci{\textbf{58.79}}{0.27} & \scoreci{61.75}{0.45} & 86M \\
        Two students & \scoreci{\textbf{65.41}}{0.14} &
        \scoreci{57.98}{0.89} & \scoreci{\textbf{62.11}}{0.37} & 172M \\
        \bottomrule
    \end{tabular}
\end{table}

Adding the DINOv3 branch to TerraMind-KD reduces the MS mean by 1.29 points,
showing modest interference from sharing capacity. On HR tasks, TEDI is 0.81
points above DINOv3-KD. Three runs are insufficient to establish a precise
positive-transfer gain, but they show that the HR capability is retained rather
than erased by sharing. Two independent students reach 62.11 overall, only
0.36 above TEDI, but require two 86M-parameter backbones. TEDI therefore
captures nearly all of the routed two-student performance with half the
backbone storage. Because each expert is tied to its corpus, this ablation does
not separate teacher knowledge from added data diversity.

\subsection{Allocating capacity to token width}
\label{sec:width_ablation}

The standard TEDI-Base trunk uses a 768-dimensional token embedding, whereas
both teachers produce 1024-dimensional targets. This motivates the hypothesis
that student token width may constitute a bottleneck: a wider shared
representation could preserve complementary information from the two teachers
more effectively. We therefore construct TEDI-B$_{1024}$ by increasing the
token dimension to 1024 while retaining 12 Transformer blocks. We increase the 
number of heads per MHA Layer from 12 to 16 in order to keep the head dimension at 64.
To limit parameter growth, we reduce the MLP expansion ratio from 4 to 1.5, reallocating
capacity from the feed-forward sublayers to the shared token representation
rather than uniformly scaling the backbone. As a control, we apply the same
architectural modification to our TerraMind-distilled and DINOv3-distilled Base models, yielding
TerraMind-KD-B$_{1024}$ and DINOv3-KD-B$_{1024}$. This comparison tests whether the modification
generally produces a stronger Base architecture or specifically alleviates the
capacity-sharing bottleneck introduced by multi-teacher distillation.

\begin{table}[t]
    \centering
    \caption{\textbf{Width-allocation ablation.} Avg. PANGAEA-9 test mIoU
    across three downstream runs, with one sample standard deviation. The B$_{1024}$ variants widen the token embedding while limiting parameter growth.}
    \label{tab:width_ablation}
    \small
    \setlength{\tabcolsep}{1.5pt}
    \begin{tabular}{lrrrrr}
        \toprule
        \textbf{Student} & \textbf{Dim.} & \textbf{Depth} &
        \textbf{MLP} & \textbf{Params.} & \textbf{All} \\
        \midrule
        TEDI-B & 768 & 12 & $4.0\times$ & 86M & \scoreci{61.75}{0.45} \\
        TEDI-B$_{1024}$ & 1024 & 12 & $1.5\times$ & 90M & \scoreci{62.32}{0.13} \\
        TEDI-L & 1024 & 24 & $4.0\times$ & 303M & \scoreci{\textbf{62.51}}{0.83} \\
        \midrule
        TerraMind-KD-B & 768 & 12 & $4.0\times$ & 86M & \scoreci{\textbf{60.69}}{0.13} \\
        TerraMind-KD-B$_{1024}$ & 1024 & 12 & $1.5\times$ & 90M & \scoreci{60.60}{0.18} \\
        \midrule
        DINOv3-KD-B & 768 & 12 & $4.0\times$ & 86M & \scoreci{57.55}{0.54} \\
        DINOv3-KD-B$_{1024}$ & 1024 & 12 & $1.5\times$ & 90M & \scoreci{\textbf{57.80}}{0.46} \\
        \bottomrule
    \end{tabular}
\end{table}

TEDI-B$_{1024}$ improves upon TEDI-B by 0.57 points while adding only 4M
parameters. It remains only 0.19 points below TEDI-L despite using 70\% fewer
parameters and 58\% fewer standardized GFLOPs, thereby closing 75\% of the
observed gap between TEDI-B and TEDI-L. In contrast,
TerraMind-KD-B$_{1024}$ and DINOv3-KD-B$_{1024}$  perform comparably to their \textit{Base} counterparts, 
without any significant score improvement. These results suggest that the benefit of the wider token
representation is specific to the multi-teacher setting, consistent with it
alleviating a capacity-sharing bottleneck rather than merely providing a
generally stronger parameter allocation.

\section{Discussion and Limitations}

\noindent\textbf{Implications for GeoFM development.}
Distillation is now a standard strategy for deriving compact model families in
general computer vision~\cite{oquab2023dinov2,simeoni2026dinov,ren2023tinymim}.
Our results show that the same principle transfers effectively to GeoFMs. CROMA 
directly compares routes with matched student architecture and corpus, showing 
that reuse of Large yields a stronger Base. The TerraMind comparison is complementary
but not fully isolating: its students both dedicate their capacity to Sentinel-2 and replace the native positional encoding and gated-MLP implementations. Their gains at 
three scales are consistent with the CROMA result, but cannot be attributed to
distillation alone. TEDI extends the same reuse principle across modalities by
consolidating two independently pretrained experts in one shared trunk.

\noindent\textbf{Unpaired expert consolidation.}
TEDI learns from independent Sentinel-2 and HR corpora, whereas paired
cross-modal methods require co-registered observations. Each expert can
therefore retain its native channels, resolution, and representative image
source without requiring a new paired dataset. This makes expert consolidation
practical even when the original pretraining corpora differ, and provides a
path for incorporating further specialist checkpoints. At deployment,
modality-specific embeddings route both input regimes through one shared
trunk, retaining most of the specialists' transfer performance with much lower
storage than deploying them independently.

\noindent\textbf{Experimental scope.}
PANGAEA-9 is a fixed subset, not the full benchmark, and equally weighted mean
mIoU can hide task-specific regressions. We study two experts and two input
regimes; SAR-only, temporal, and broader modality combinations remain open.
The teacher-multiplicity ablation adds a second image corpus together with the
second expert, so it measures their practical joint contribution rather than
separating teacher knowledge from additional data diversity. Finally, three
downstream runs limit conclusions about very small differences between nearby
operating points.

\noindent\textbf{Data and release.}
Distillation uses the public SSL4EO-S12 and FMoW corpora
\cite{wang2023ssl4eo,christie2018fmow}; no source imagery is redistributed.
Both provide broad geographic coverage, but FMoW emphasizes functional sites
and human infrastructure, and neither represents every sensor, temporal
regime, or sparsely observed region. TEDI also inherits its teachers' biases.
Upon acceptance, we plan to release training and evaluation code, model cards,
preprocessing details, and eligible checkpoints. Code will use a permissive
license; checkpoint distribution will preserve the upstream CROMA, TerraMind,
and DINOv3 terms~\cite{fuller2023croma,jakubik2025terramind,simeoni2026dinov}.

\section{Conclusion}

We studied feature distillation as a common mechanism for two practical GeoFM
problems: deriving efficient backbone families and consolidating
modality-specific experts. A frozen large teacher produces Tiny, Small, and
Base students that improve the released TerraMind accuracy--compute frontier.
The multi-expert extension, TEDI, uses independent Sentinel-2 and HR corpora to
combine frozen TerraMind and DINOv3 teachers in one shared backbone. On
PANGAEA-9, TEDI-Large achieves the highest single-model average and
matches both specialist domain means, while TEDI-Base retains most of this
performance with a substantially smaller backbone. Reallocating Base capacity
toward a 1024-dimensional token space further closes most of the remaining gap
to TEDI-Large with only a modest parameter increase. These findings support a
simple design principle for future GeoFM releases: pretrain strong experts
once, then reuse their representations across the required model scales and
modalities.

\section*{Acknowledgments}
This work was funded by the French Research Agency (ANR) through the DREAMS (ANR-25-CE56-4384-01) project, by the French Space Agency (CNES), and was provided with computing HPC and storage resources by GENCI at IDRIS thanks to the grant 2025-AD011015819R1 on the supercomputer Jean Zay’s V100 and H100 partitions.

{
    \small
    \bibliographystyle{ieeenat_fullname}
    \bibliography{main}
}

\end{document}